\documentclass[letterpaper, 10 pt, conference]{ieeeconf}  

\IEEEoverridecommandlockouts                              

\usepackage{amsmath} 
\usepackage{amssymb}  
\usepackage{graphicx}
\usepackage{algorithm}
\usepackage{algpseudocode}
\usepackage{booktabs}
\usepackage{algorithm}
\usepackage{float}
\usepackage{stfloats}

\title{\LARGE \bf

A Transferable Legged Mobile Manipulation Framework Based on Disturbance Predictive Control
}

\author{Qingfeng Yao$^{1,2,3*}$, Jilong Wang$^{4*}$, Shuyu Yang$^{4*}$, Cong Wang$^{1,2,3}$, Linghan Meng$^{1,2,3}$ \\
Qifeng Zhang$^{1,3}$, Donglin Wang$^{4+}$
\thanks{* Contributed equally}
\thanks{+ Corresponding author. Email: wangdonglin@westlake.edu.cn}%
\thanks{The main work was done at MiLAB, Westlake University.}
\thanks{$^{1}$State Key Laboratory of Robotics, Shenyang Institute of Automation, Chinese Academy of Sciences, Shenyang 110016, China}%
\thanks{$^{2}$Institutes for Robotics and Intelligent Manufacturing, Chinese Academy of Sciences, Shenyang 110169, China}%
\thanks{$^{3}$University of Chinese Academy of Sciences, Beijing 100049, China}%
\thanks{$^{4}$School of Engineering, Westlake University, Hangzhou 310024, China}%
}

\begin{document}

\maketitle
\thispagestyle{empty}
\pagestyle{empty}

\begin{abstract}

    Due to their ability to adapt to different terrains, quadruped robots have drawn much attention in the research field of robot learning. 
    Legged mobile manipulation, where a quadruped robot is equipped with a robotic arm, can greatly enhance the performance of the robot in diverse manipulation tasks. 
    Several prior works have investigated legged mobile manipulation from the viewpoint of control theory. 
    However, modeling a unified structure for various robotic arms and quadruped robots is a challenging task.
    In this paper, we propose a unified framework disturbance predictive control where a reinforcement learning scheme with a latent dynamic adapter is embedded into our proposed low-level controller. 
    Our method can adapt well to various types of robotic arms with a few random motion samples and the experimental results demonstrate the effectiveness of our method.  
\normalcolor

\end{abstract}


\section{Introduction}

Legged robots have shown great advantages when facing challenging terrain compared over other types of robots due to their unique locomotion pattern. This advantage is achieved by adjusting the ground reaction force, which allows the robot to maintain balance and achieve compliant behavior. Traditional autonomous robots contain three components: navigation, motion planning and control~\cite{bledt2017policy,dudzik2020robust}. 
A high-level model has been established to analyze the environment and to generate the desired robotic motion, and a low-level controller has been used to ensure that the desired action is executed correctly~\cite{grandia2020multi,jenelten2020perceptive}. However, traditional model-based control methods depend highly on the accuracy of the dynamic model, so the control effectiveness decreases with the increasing discrepancy between the actual system and the mathematical model.

A legged robot is capable of undertaking complex tasks including search, rescue, exploration and inspection when equipped with a robotic arm. In such tasks, manipulation ability becomes very important. Although it is still possible to use gripper-like feet~\cite{heppner2014versatile} to complete possible manipulation tasks during such missions, using an independent manipulator achieves better stability~\cite{rehman2016towards,abe2013dynamic}.
There are several challenges in a legged mobile manipulation system. It requires proper motion planning and control to be able to move and manipulate at the same time~\cite{ewen2021generating}.
This whole-body dynamic interaction with the environment needs to consider the dynamics of the entire system and the influence of the manipulator of the robot. Traditional control methods have several limitations such as a lack of adaptability.
\\
At the same time, reinforcement learning, a method that has emerged rapidly in machine learning in recent years, has gradually shown its great potential in decision-making problems. Deep reinforcement learning (DRL) explores the environment by interacting with the environment and utilizing a neural network~\cite{mnih2015human}, and it adjusts its strategies according to the reward to learn from scratch and complete the given tasks. In the case of quadruped robot control, DRL reduces the need for accurately modeling the virtual environment and 
enables the robot to adapt to a new environment. It is feasible to solve different robot control problems using DRL and important breakthroughs have been made in quadruped robots~\cite{hwangbo2019learning,yang2020multi}. 


To solve the modeling difficulty and poor portability problems of legged-arm robots, 
in this paper, we further propose a framework for integrating a robotic arm and a quadruped robot, where the latent influence of external limbs is predicted via a migratable forward model. DRL outputs the disturbance coefficient to the low-level controller to help keep the body steady with latent influence.
In summary, this work made the following contributions:

\begin{figure}[tbp]
\centering
\includegraphics[width=0.4\textwidth]{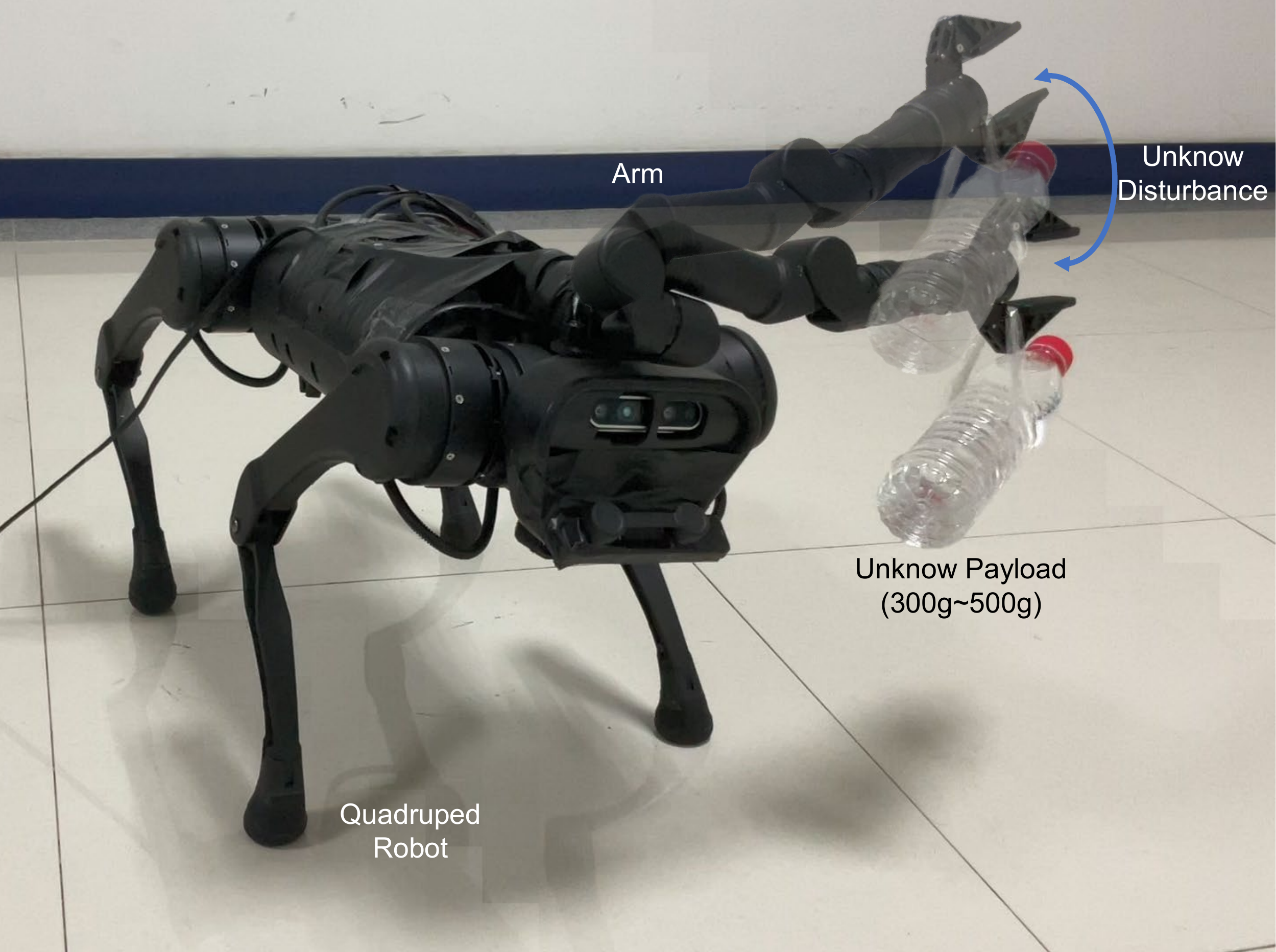}
\caption{
The legged mobile manipulation with unknown disturbance.
}
\label{Fig.robot}
\end{figure}


\begin{itemize}%
\item We are using DRL to solve the control collaboration problem between robotic arms and a quadruped robot. A low-level optimal control model is proposed. By using the effect of manipulation as the input of the control model, we grant the model the ability to eliminate the influence of the robotic arm with an appropriate disturbance parameter. 
\item High-level reinforcement learning estimator is used to estimate the appropriate coefficient with a transferable forward model.
\item Experiments with different forms of robotic arms demonstrate the transferability of our algorithm.
\end{itemize} 


\section{Related Works}
\label{sec:citations}
\subsection{Control-based Legged Locomotion}
	Mainstream model-based control methods of quadruped robots utilize the model predictive control (MPC) method to simplify the whole-body control problem to an optimization problem~\cite{grandia2020multi,di2018dynamic,bellicoso2018dynamic} and further complete various tasks, such as walking up stairs~\cite{jenelten2020perceptive,villarreal2020mpc,gangapurwala2020guided,fankhauser2018probabilistic}.
	Mobile manipulators expand the ability of robots to interact with their environment~\cite{minniti2019whole}.  
	Recent work has investigated a mobile quadrupedal control framework with robotic arms that can achieve dynamic gaits while performing manipulation tasks. 
	The robot responds submissively to external forces and maintains balance while performing dynamic movements~\cite{bellicoso2019alma}.
	An alternative approach is to construct a holistic MPC framework that plans the entire body movement/force trajectory according to the task of dynamic movement and manipulation~\cite{sleiman2021unified}.

    We propose a hierarchical robot control method to combine the benefits of reinforcement learning and optimization methods. To incorporate the interference caused by the robotic arm to the quadruped robot, we model the dynamics of the quadruped robot considering the disturbance of the manipulator. 
    The control method is used to maintain body balance with distraction estimated by DRL. This structure decouples the body and the arm to improve the adaptability of the body to different robotic arms.
\subsection{Learning-based Legged Locomotion}
    In recent years there have been many applications of applied reinforcement learning to robot control problems, especially those related to quadruped robots~\cite{tsounis2020deepgait,lee2020learning,lee2020learning}.
    A quadruped robot can learn sensor information and apply it to multiple skills with reinforcement learning~\cite{escontrela2020zero,bellegarda2020robust,hoeller2021learning,jain2020pixels,paigwar2020robust}.
    To enhance the efficiency of reinforcement learning, several hierarchical networks~\cite{gangapurwala2020rloc,da2020learning} combine control methods with deep learning to merge the advantages of both. With DRL being responsible for the high-level estimator of inputting commands into the underlying control methods, the hierarchical structure can complete the task in both an adaptive and stable manner.

    It is challenging to rely entirely on reinforcement learning to manipulate a legged mobile robot. In this paper, DRL is used to assist the optimizer in maintaining body stability. It does so without expert knowledge by inferring the dynamic parameters based on the robot's current latent state.

\subsection{Forward Model Learning}
    To better aid reinforcement learning for exploration, a dynamic model of the learning environment is a viable idea~\cite{henaff2019model,ha2018world}.
    A world model is learned to train agents through imagined self-play to achieve agility in action sequences~\cite{schwarting2021deep,nair2017combining,pathak2018zero}, and the latent strategies of opponents can be predicted to guide the agent toward appropriate directions~\cite{xie2020learning}.
    Dynamic model learning also plays a very important role in curiosity~\cite{burda2018large,pathak2017curiosity}.
    The curiosity-based model establishes a relationship between state change and the action of the agent by building a dynamic model.
    Forward and inverse dynamics are proposed to learn in an unsupervised manner for contrastive estimation~\cite{wang2020cloud}.

    We combine a forward model with reinforcement learning. The forward model extracts the latent state according to the current state for DRL to estimate the distractions experienced by various robotic arms at the moment.


\section{Methodology}
\label{sec:method}

\subsection{Disturbance Predictive Control}
\begin{figure*}[tbp]
    \centering
    \includegraphics[width=0.8\textwidth]{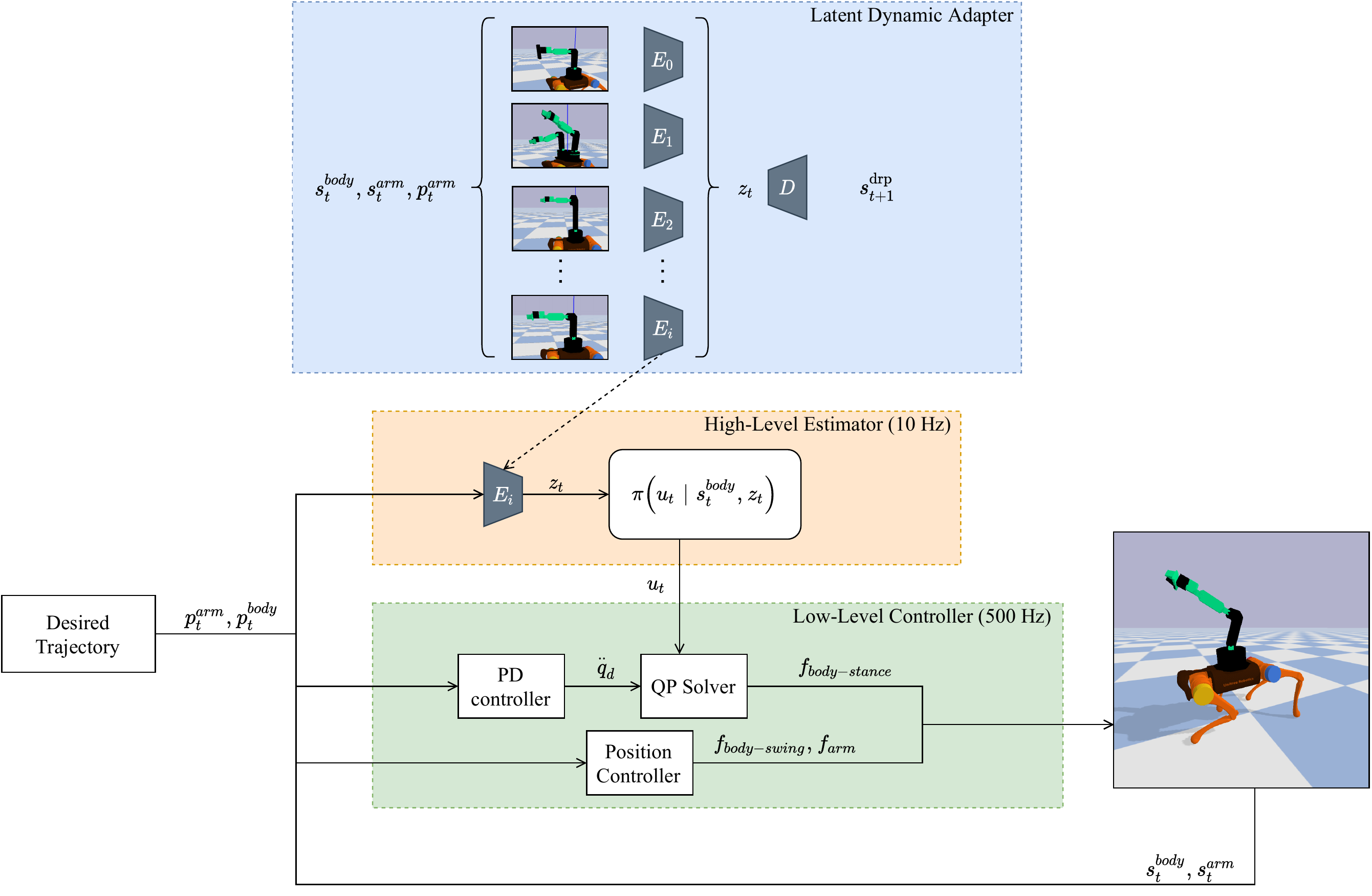}
    \caption{
    Overview of our method. 
    The first part is a latent dynamic adapter based on the forward model. 
    It estimates the latent state of the manipulator and has the ability to quickly migrate to different types of manipulator.  
    The second part is a high-level estimator based on DRL. 
    It receives the latent dynamics state of the robot from the latent dynamic adapter and sends disturbance dynamic parameters to the low-level controller. 
    The last part is an improved control method as the low-level controller. It executes the desired motion and maintains the balance of the robotic arm based on disturbance dynamic parameters.}
    \label{Fig.overview}
    \end{figure*}

    To build a framework that is adaptable to changes in the dynamics of the manipulation, we propose a framework to reduce the disturbances caused by different robotic arms. Our method isolates the dynamic impact of manipulation from the whole-body system, latent dynamic adapter estimates the latent state of different robotic arms and a DRL model predicts dynamic parameters with the latent state extracted to the low-level controller as shown in Fig. \ref{Fig.overview}.

    The robot only knows the state of the body and the arm without the information about the force applied to the arm or body. It can be modeled as a partially observable Markov decision process(POMDP), defined by a tuple,
    $\langle \mathcal{S}, \mathcal{U}, P, \mathcal{R}, \Omega, O, \gamma\rangle$,
    where $s_t \in \mathcal{S}$ describes the state of the environment, the agent chooses action $u_{t} \in \mathcal{U}$, and next state $s_{t+1} \sim P\left(s_{t}, u_{t}\right)$ is calculated by the transition kernel $P$ and the cooperative reward $r_{t}=\mathcal{R}\left(s_{t}, u_{t}\right)$ is returned to the agent.

    The full state of the environment is hidden from the agent, and the agent obtains its observation $o_{t} \in \Omega$ from an observation kernel $O\left(s_t, u_t\right)$. The agent chooses an action based on its individual history.
    The agent aims to learn a policy to maximize the total discounted reward $\mathbb{E}\left[\sum_{t=0}^{\infty} \gamma^t r_{t}\right]$,
    where $\gamma \in[0,1)$ is a discount factor.



\subsection{Latent Dynamic Adapter}
    The robotic arm continuously disturbance the legged robot and will cause the body to shake and further lose balance. To learn the disturbance of the body for different situations of the robotic arm, we introduce a forward model called latent dynamic adapter to predict the latent dynamic impact of the robotic arm on the body.
    The latent dynamic adapter predicts the roll rate and pitch rate $s_{t+1}^{drp}$ of the body at the next moment based on the current state of the body $s_{t}^{body}$ and the state of arm $s_{t}^{arm}$ and the desired joint position $p_t^{arm}$ of the manipulation which calculated from the reference motion. The purpose of this operation is to relate the state and position of the robotic arm to the changes in the body state. The state $s_{t}^{body}$ contains root orientation (read from the IMU) and the state $s_{t}^{arm}$ contains joint angles of the arm. The model is built to learn the relationship between the manipulation and the change of the body state and obtains the current influence of the manipulator through the latent state $z_t$. The latent state $z_t$ in this article is a 2-dimensional vector.  
    
    The influence of the robotic arm can be represented by the latent state. The latent state will change if the robotic arm's position changes or the arm is forced. To assist the robot resist disturbances of the robotic arm, the high-level estimator needs learn to use this latent state to output appropriate dynamic parameters.

    In practice, the latent dynamic adapter is trained by the following optimization: 
    \begin{equation}
    \max _{\phi, \psi} \log p_{\phi, \psi}\left(s_{t+1}^{drp} \mid s_{t}^{body},s_{t}^{arm}, p_{t}^{arm}\right)
    \end{equation}
    where $\mathcal{E}_{\phi}$ is a encoder and $\mathcal{D}_{\psi}$ is a decoder. $z_t=\mathcal{E}_{\phi}\left(s_{t}^{body},s_{t}^{arm}, p_{t}^{arm}\right)$ obtains the latent state of time $t$ by the encoder. $s_{t+1}^{drp}$ is estimated by the decoder with the latent state $z_t$. 
    To train the latent dynamic adapter, data are collected based on the random movement of both the robot and the arm. The relationship between change of state of the quadruped robot and various robotic arms can be predicted using the trained latent dynamic adapter.

    

\subsection{High-Level Estimator}

The high-level estimator provides the dynamic parameters $u_t = \pi_{\theta}\left(s_t^{body}, z_t\right)$ to the low-level controller using the current body state $s_{t}^{body}$ and latent state $z_t$. 
The goal of high-level estimator is to estimate the control model's dynamics parameter and enable the quadruped and robotic arm to follow the given trajectory of different tasks. The reward function is designed as the error with the desired trajectory $p_t^{body}$:
    \begin{equation}
    \begin{aligned}
    r_{vel}:=\exp (-8.0(v_{x} -v_{x^d})^{2}) + \exp (-8.0(v_{y} -v_{y^d})^{2}) \\ + \exp (-8.0(\omega_{yaw} -\omega_{yaw^d})^{2})
    \end{aligned}
    \end{equation}
    \begin{equation}
    \begin{aligned}
    r_{orn}:=\exp (-8.0(\theta_{roll} - \theta_{roll^d})^{2}) \\ + \exp (-8.0(\theta_{pitch} - \theta_{pitch^d})^{2})
    \end{aligned}
    \end{equation}
    \begin{equation}
    r_{total}:=0.08*r_{vel} + 0.05*r_{orn}
    \end{equation}
    where $r_{vel}$ is the speed tracking reward, $v_{x}$, $v_{y}$ and $\omega_{yaw}$ are the current linear velocity and angular velocity of the body, and $v_{x^d}$, $v_{y^d}$ and $\omega_{yaw^d}$ are the desired linear velocity and angular velocity of the body. $r_{orn}$ is the angle tracking reward, $\theta_{roll}$ and $\theta_{pitch}$ are the current roll and pitch of the body, respectively, $\theta_{roll^d}$ is the desired roll, $\theta_{pitch^d}$ is the desired pitch and $r_{total}$ is the total reward.
    
    The trained high-level estimator output is based on the body state and the latent state, which means a trained estimator can be used for different robotic arms with similar latent states.
    The trained decoder can map the latent state $z_t$ to the $s_{t+1}^{drp}$ and we use the trained decoder as a supervisor for new encoders.  
    When a new robotic arm is employed, the new encoder can be trained with the trained decoder, and the new encoder will maps the input to a similar latent state via the decoder's constraint. 
    This means that the original high-level estimator can use the latent state directly from the new encoder.
    To complete the strategy migration, the deployment of a new manipulator only needs to collect random motion to train the encoder.
\subsection{Low-Level Controller}
We use a model-based control method to accomplish quadruped robot locomotion tasks. The high-level RL framework will provide the estimated dynamic parameters $u_t = [f_a, \tau_a]^{T}$. Then the base pose controller will compute the ground reaction force of the standing leg to track the desired base posture. We also use standard proportional-derivative (PD) control for the leg swing to follow the desired trajectory. This structure optimizes control and maintains balance with different robotic arms.

We approximate the quadruped dynamics as linearized rigid body dynamics based on~\cite{di2018dynamic} and further consider the influence of the robotic arm:

\begin{equation}
\ddot{q}=\mathbf{M} f-\tilde{g} + \mathbf{A} f_{a} + \mathbf{B} \tau_{a}
\end{equation}

\begin{equation}
\mathbf{M}=
\begin{bmatrix}
\frac{1_3}{m} & \frac{1_3}{m} & \frac{1_3}{m} & \frac{1_3}{m} \\
\mathbf{R}_z^\intercal I_{B}^{-1} [r_1] & \mathbf{R}_z^\intercal I_{B}^{-1} [r_2]  & \mathbf{R}_z^\intercal I_{B}^{-1} [r_3] & \mathbf{R}_z^\intercal I_{B}^{-1} [r_4]
\end{bmatrix}
\end{equation}

\begin{equation}
\mathbf{A}=
\begin{bmatrix}
\frac{1_3}{m} \\
0_3
\end{bmatrix}
\end{equation}

\begin{equation}
\mathbf{B}=
\begin{bmatrix}
0_3\\
\mathbf{R}_z^\intercal I_{B}^{-1}
\end{bmatrix}
\end{equation}

where $\ddot{q}$ is the acceleration of body and $\mathbf{M} \in \mathbb{R}^{6 \times 12}$ is the inverse inertia matrix and $f=\left(f_{1}, f_{2}, f_{3}, f_{4}\right) \in \mathbb{R}^{12}$ is the vector of the Cartesian forces of the four feet.
$\tilde{g}=\left(g, 0_{3}\right) \in \mathbb{R}^{6}$ is the gravity vector. $f_{a} \in \mathbb{R}^{6}$ and $\tau_{a} \in \mathbb{R}^{6}$ are reaction force and reaction torque from the robotic arm, estimated by the RL network. We connect the optimal control and DRL through this parameter. $\mathbf{A} \in \mathbb{R}^{6 \times 3}$ is the inverse inertia matrix for the robotic arm reaction force $f_{a}$. $\mathbf{B} \in \mathbb{R}^{6 \times 3}$ is the inverse inertia matrix for the robotic arm reaction torque $\tau_{a}$.
From equation (6) to (8),
$I_{B}$ is the base inertia.
$R_{z}^\intercal$ is the rotation matrix of yaw. $r_i=( r_1, r_2, r_3, r_4 )$ are the vector from the center of mass (COM) to the point where the force is applied.
m is the mass of the robot.

Note that $f_{a}$ and $\tau_{a}$ represent the effect of the arm predicted by a high-level estimator. The desired base pose $q_{d}$ and velocity $\dot{q}_{d}$ provided by the desired trajectory, and the target acceleration is solved as follows:
\begin{equation}
\ddot{q}_{d}=k_{p}\left(q_{d}-q\right)+k_{d}\left(\dot{q}_{d}-\dot{q}\right).
\end{equation}
The optimal control based on disturbance is established as follows:
\begin{equation}
\begin{aligned}\label{MPC}
&\min _{f}\left\|\mathbf{M} f-\tilde{g}-\ddot{q}_{d}+\mathbf{A} f_{a} + \mathbf{B} \tau_{a}\right\|_\mathbf{Q}+\|f\|_{\mathbf{R}}\\
&\text { st. } \quad f_{z, i} \geq f_{z, \min }\\
&\begin{array}{l}
-\mu f_{x} \leq f_{z} \leq \mu f_{x} \\
-\mu f_{y} \leq f_{z} \leq \mu f_{y}
\end{array},
\end{aligned}
\end{equation} \label{eq.3}
where Q and R are diagonal matrices in the cost function.

Moreover, the foot force for each foot swing is computed as follows:
\begin{equation}
f_{i}=k_{p, i}\left(p_{d, i}-p_{i}\right)-k_{d, i} \dot{p}_{i},
\end{equation} \label{eq.4}
where position $p_{d,i}$ is the desired foot position for foot $i$ that is calculated by the high-level estimator, $p_i$ is the current foot position and $\dot{p}_{i}$ is the current velocity for foot $i$.
The foot forces are converted to torques with $\tau=J^{T} f$, where $J \in \mathbb{R}^{12 \times 12}$ is the Jacobian matrix.


\section{Experimental Results}
\label{sec:results}

\begin{figure}[htbp]
    \centering
    \includegraphics[width=0.11\textwidth,height=0.1\textheight]{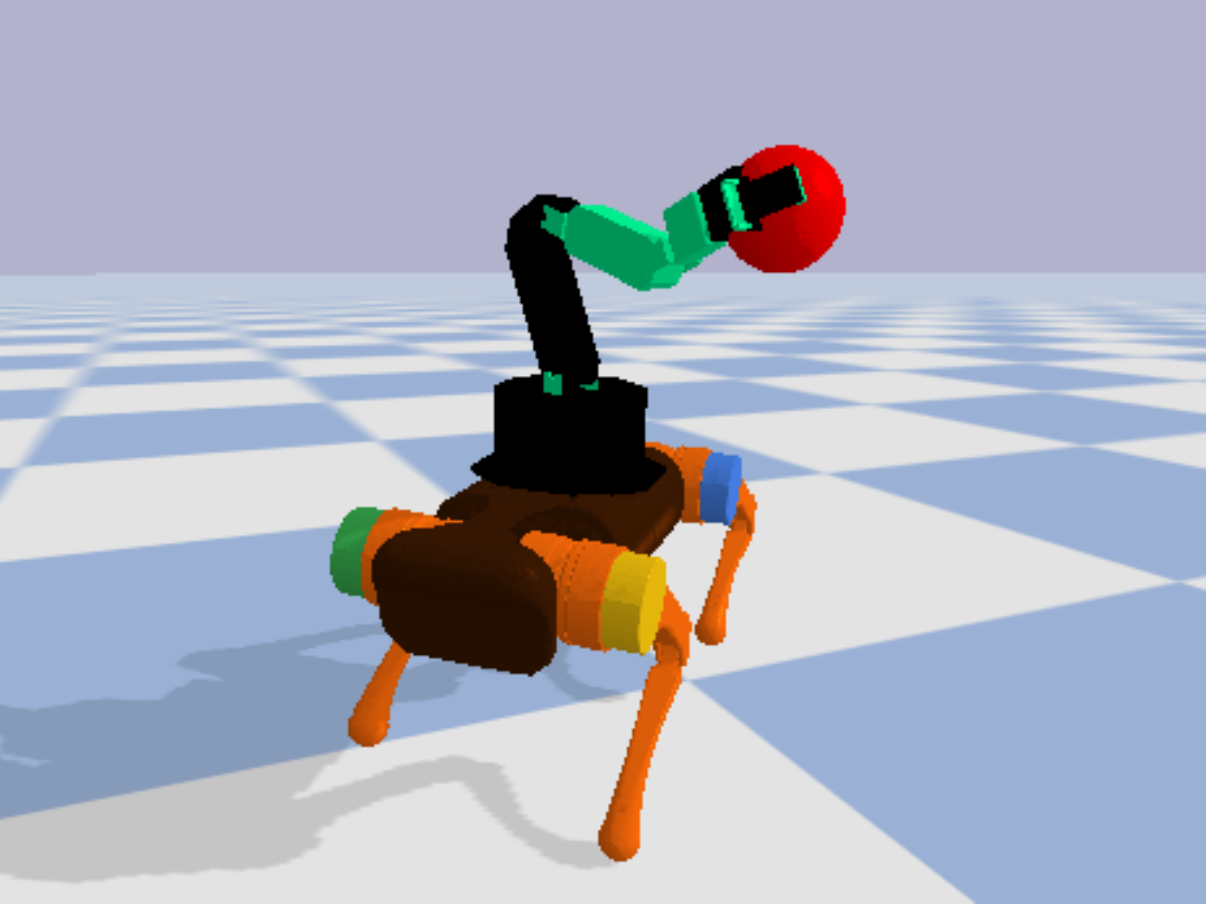}
    \includegraphics[width=0.11\textwidth,height=0.1\textheight]{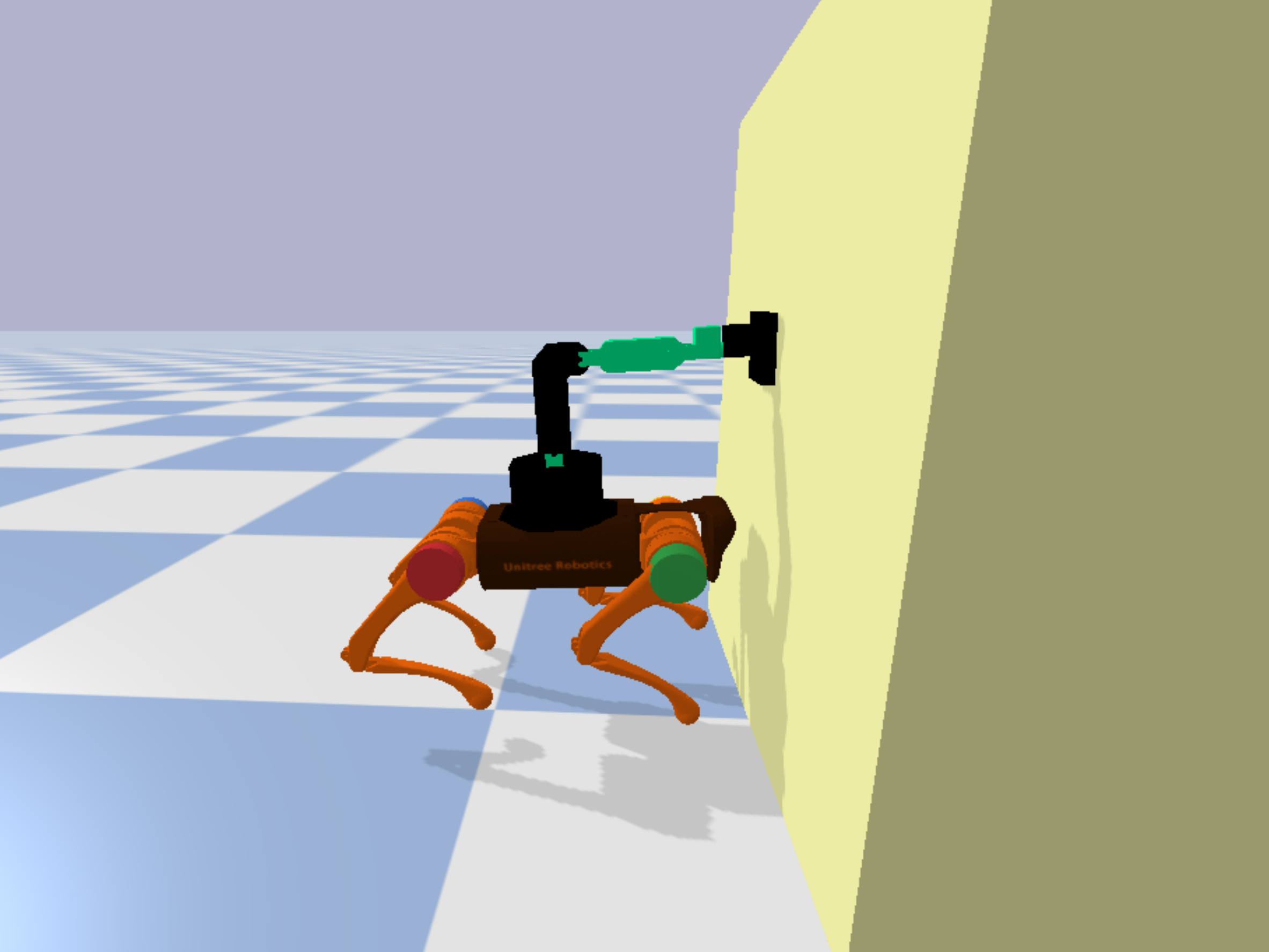}
    \includegraphics[width=0.11\textwidth,height=0.1\textheight]{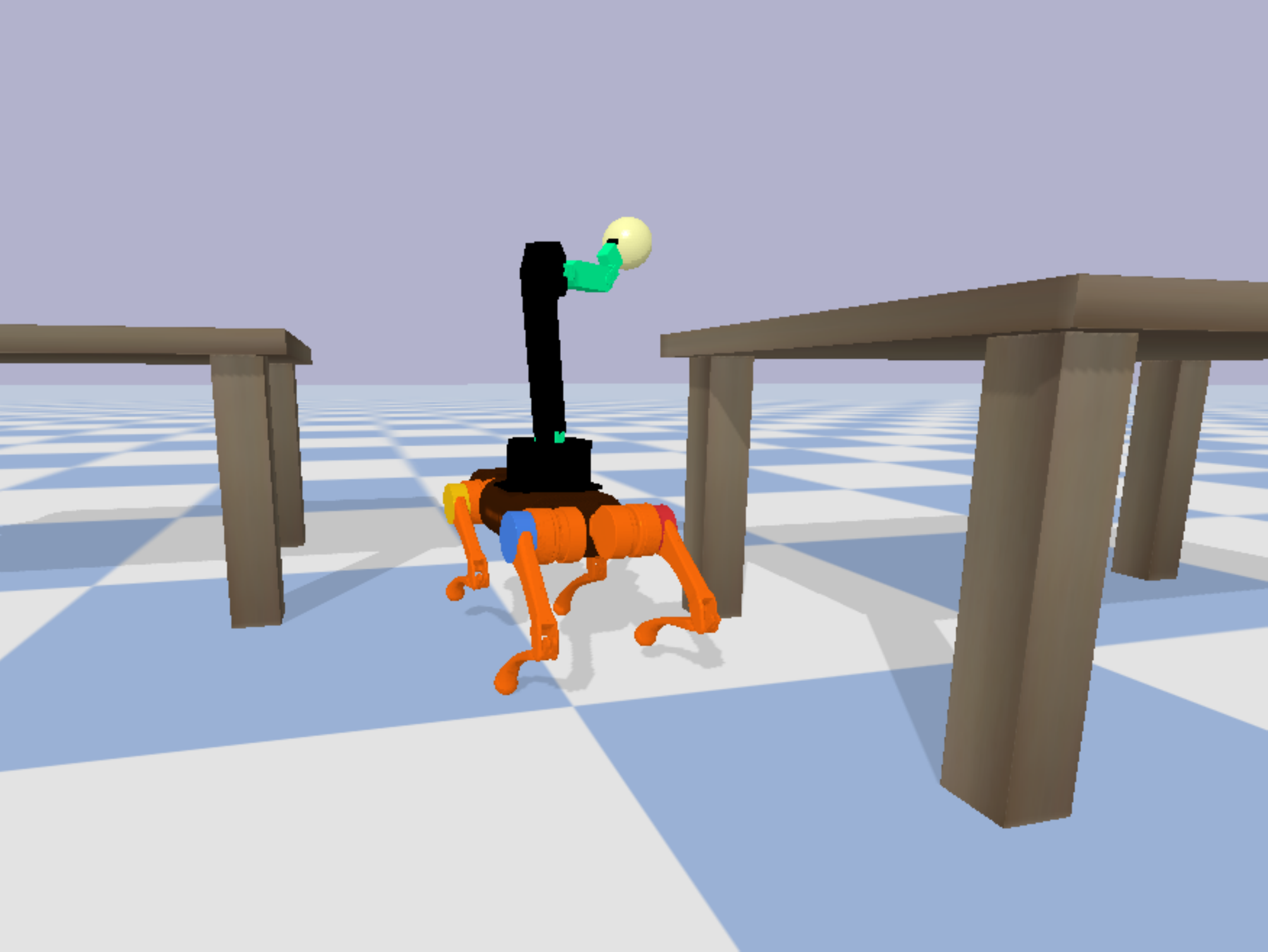}
    \includegraphics[width=0.11\textwidth,height=0.1\textheight]{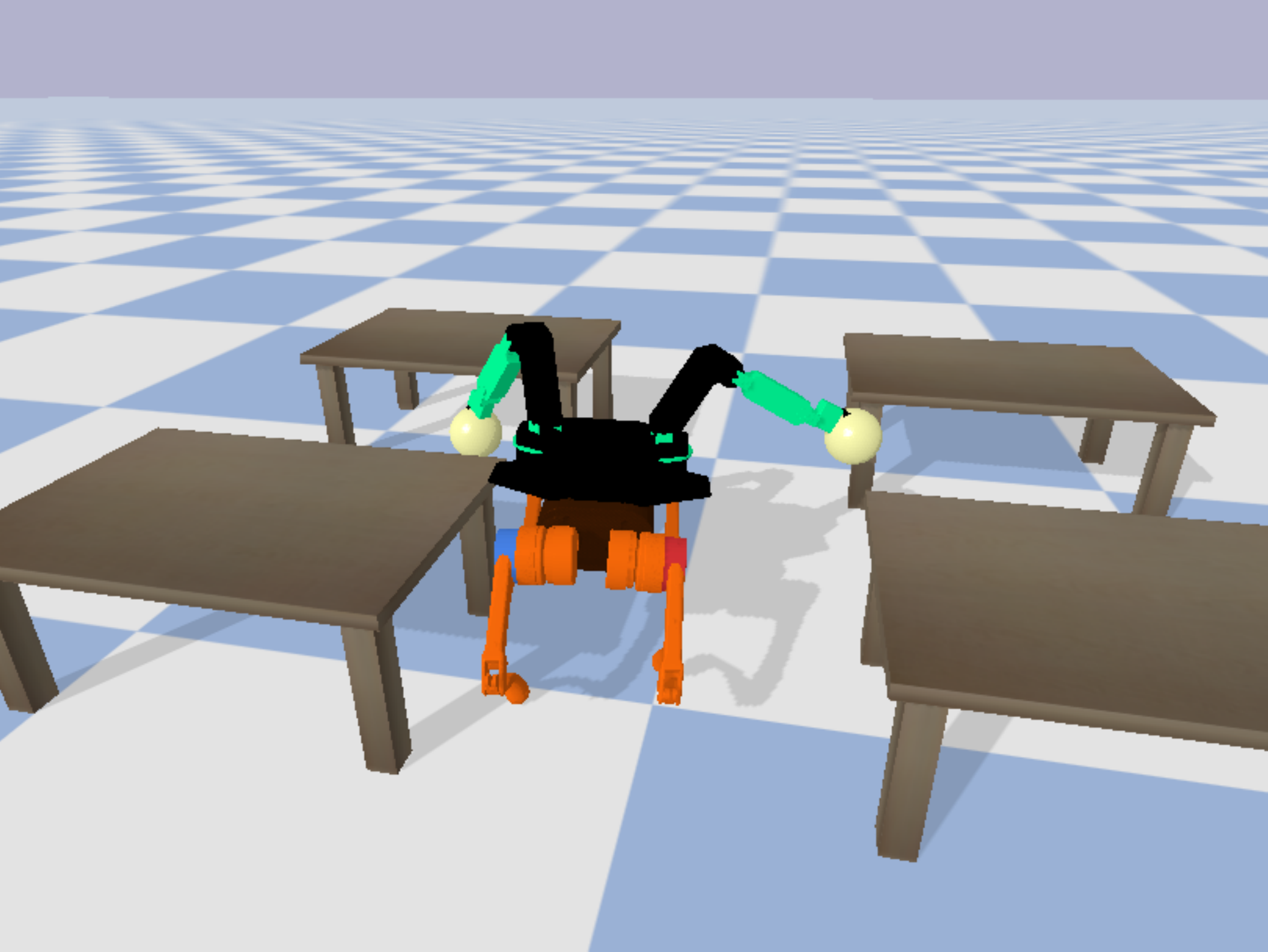}
    \caption{Sketch of the goal-reaching task, block-pushing task, and ball-carrying task}
    \label{Fig.trans}
    \end{figure}

    Our framework is used to complete three mobile robotic manipulation tasks as shown in Fig. \ref{Fig.trans}. Different robotic arms are used depending on the task. Our framework can be deployed on different types of robotic arms with little data collection.
    Several tasks and corresponding robotic arm movements are performed in the experiment:
    
    \begin{itemize}%
    \item Reaching: The task is to move the end gripper of the robotic arm to the target position. In this task, the robot needs to balance the effects of different robotic arm positions under random forces.
    The policy is trained with this task and learns to output the appropriate dynamic parameters according to the latent state and the body state. This policy can be directly used in other tasks without fine-tuning.
    \item Pushing: The task is to move a box to the target position. This task tests the ability for simultaneous movement of the body and arm.
    In this task, the robot needs to remain stable under the constant force on the robotic arm. A heavier but flat robotic arm is used for this task to help push the box steadily.
    \item Carrying: The goal is to move a ball from one table to another.
    The robot needs to maintain its steady while utilizing the robotic arm to move the ball in this experiment.
    In this task, we also explored two scenarios involving a higher table and the movement of multiple balls, which require a distinct type or number of robotic arms.
    \end{itemize} 
    
    We use a quadruped robot with a robotic arm and run the environment in PyBullet. For the quadruped robot, we pick A1, a Unitree robot, that has 12 degrees of freedom, three for each leg. Both arms of the robotic arm have four degrees of freedom (ignoring gripper joints). We make four different types of arms: a WidowX robotic arm of standard size and weight, a WidowX robotic arm with a heavier end effector, a WidowX robotic arm with a longer link length, and two normal WidowX arms.
    The robot needs to keep its body stable and move according to a given trajectory when equipped with different robotic arms.

    The policy is trained in the reaching task, and a random force is applied to the robotic arm to simulate the force on the arm in various tasks.
    The learned policy can output appropriate dynamic parameters based on the latent state from distinct encoders. The trained policy can be directly applied to different tasks and robotic arms without fine tuning because the input of policy is the latent state and the body state rather than the state of the arm.
    To complete the migration to the new manipulator, it merely needs to collect data from random movement to train the encoder supervised by a trained decoder.

    Our computer, based on Ubantu 18.04 and NVIDIA-2080 takes five hours to perform the iterations to finish training the robot. 
    The SAC algorithm~\cite{haarnoja2018soft} has 2 hidden layers for the actor and critic networks, each with 128 hidden units. The hidden layers use the ReLU activation function. It is trained by backpropagation using Adam as the optimizer. The encoder and decoder have 2 hidden layers with 128 hidden units. The ReLU activation function is used by the encoder and decoder.
    
\begin{figure}[htbp]
    \centering
    \includegraphics[width=0.23\textwidth]{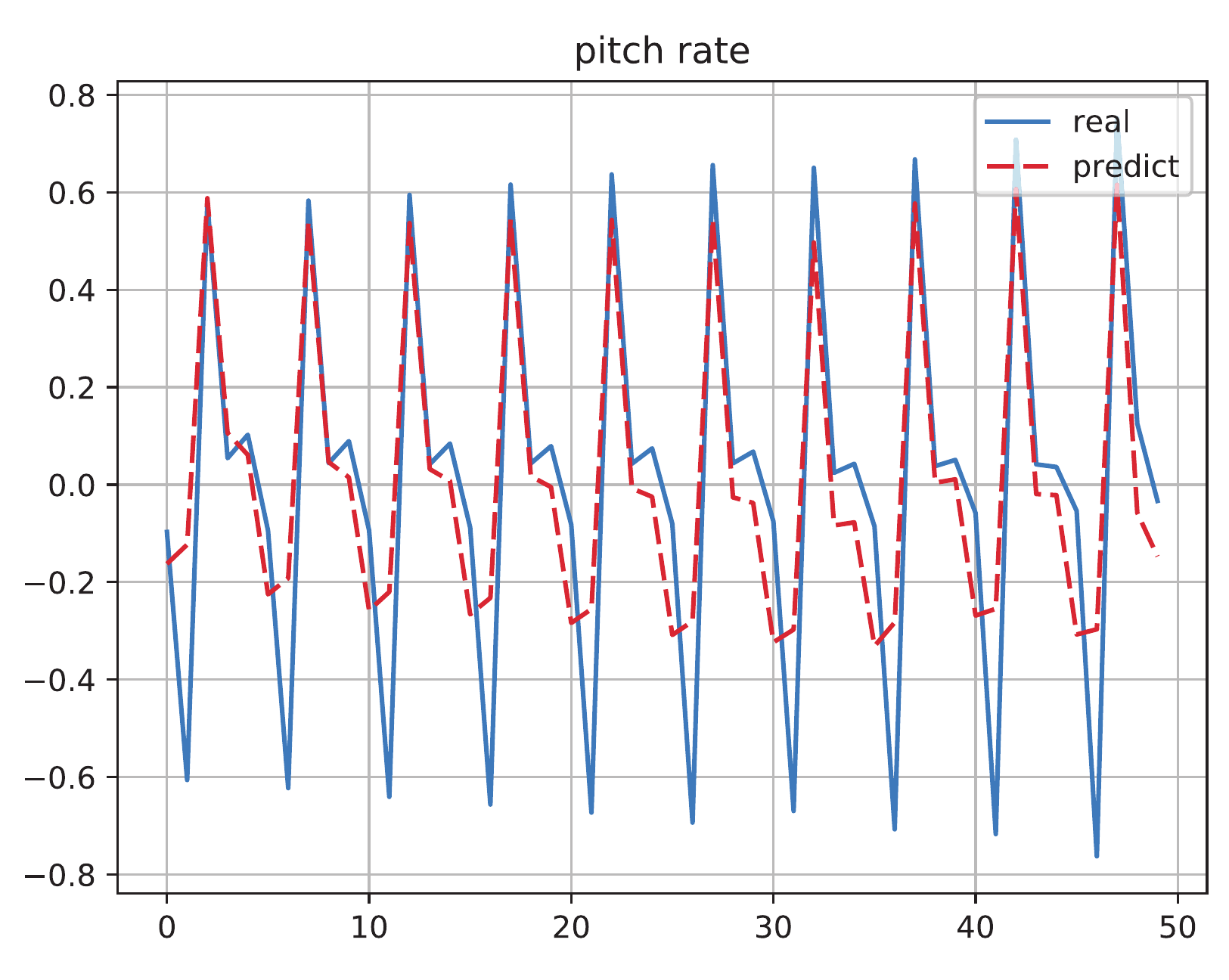}
    \includegraphics[width=0.23\textwidth]{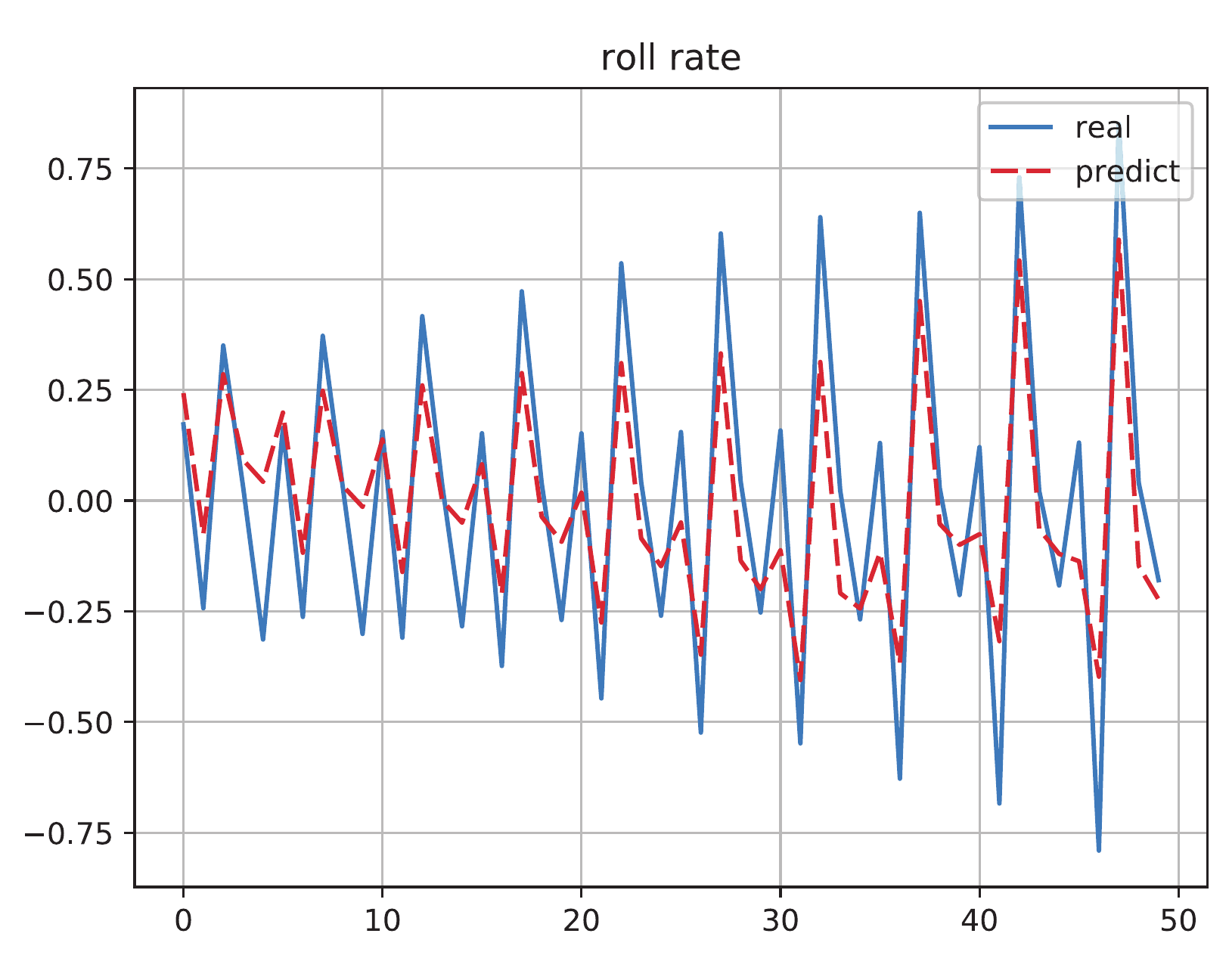}
    \caption{Comparison of the predicted posture change and real posture change. Our forward model is capable of learning how does the change of the body during the motion of the robot.}
    \label{Fig.predict}
    \end{figure}

    We test the predictive ability of the forward model in Reaching task, the results demonstrate the decoder can estimate changes in body posture based on the latent state $z_t$ as shown in Fig. \ref{Fig.predict}. 
    According to the current state of the manipulator and body, the forward model can anticipate the next state of the body, indicating that the latent state $z_t$ can represent the impact of the manipulator.

\begin{figure}[htbp]
    \centering
    \includegraphics[width=0.5\textwidth]{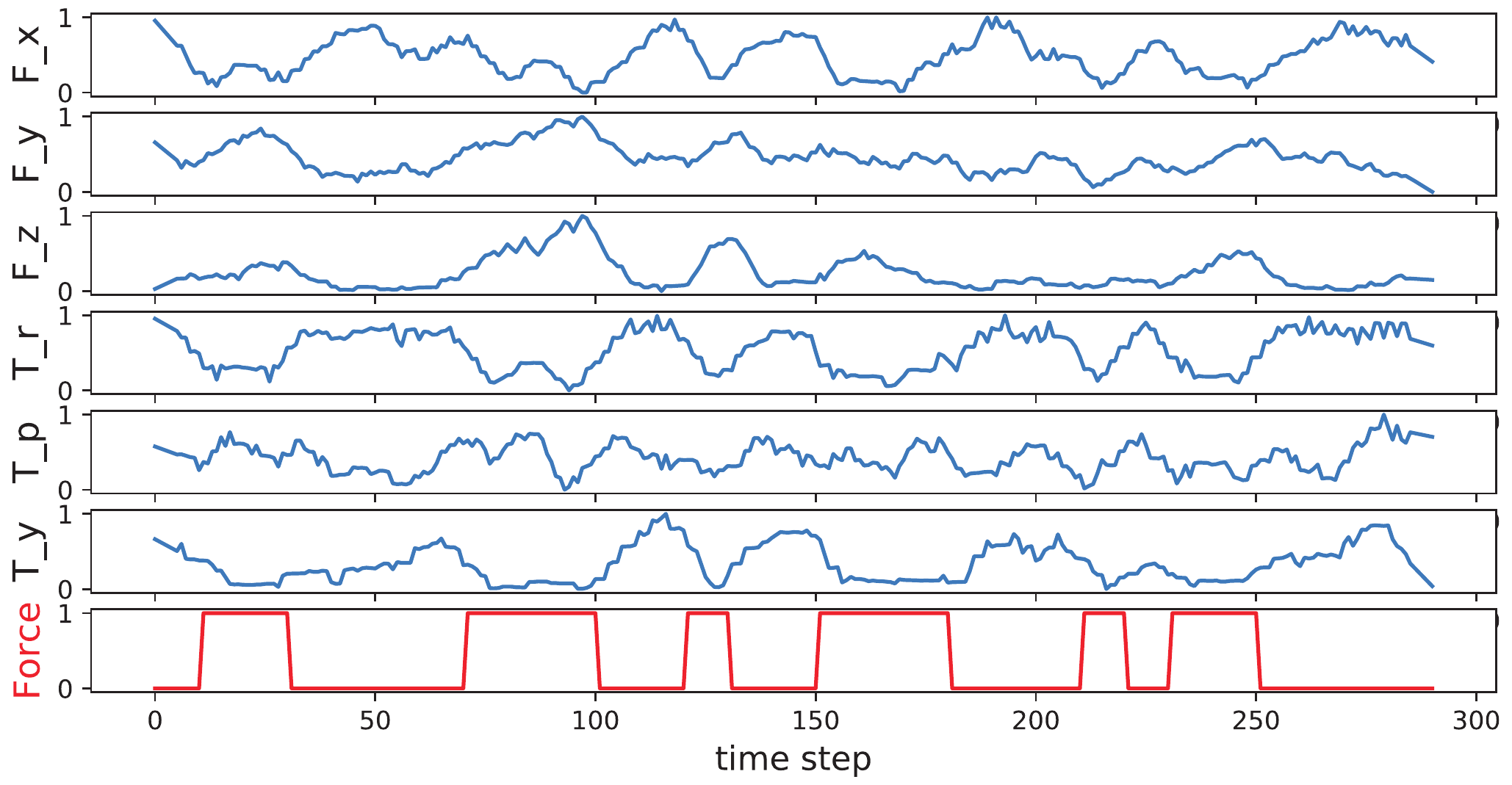}
    \caption{
    Output parameters of DRL with the sudden force applied to the robotic arm, the high-level estimator adjusts the output when the force is present.}
    \label{Fig.cmp}
    \end{figure}

    We further test the adaptability of the network under the change of the force on the arm. Despite the absence of force sensors, the high-level estimator learns to estimate the dynamically changing dynamic parameters from the latent state and body state, as shown in Fig. \ref{Fig.cmp}.
    The results reveal that, despite the lack of the force sensor, the high-level estimator changes dynamic parameters based on body state and latent perception.
    It implies the output dynamic parameters of the high-level state estimator can be corrected in response to the changes, assisting the robot in maintaining stability.
    
    \begin{figure}[htbp]
    \centering
    \includegraphics[width=0.4\textwidth,height=0.2\textheight]{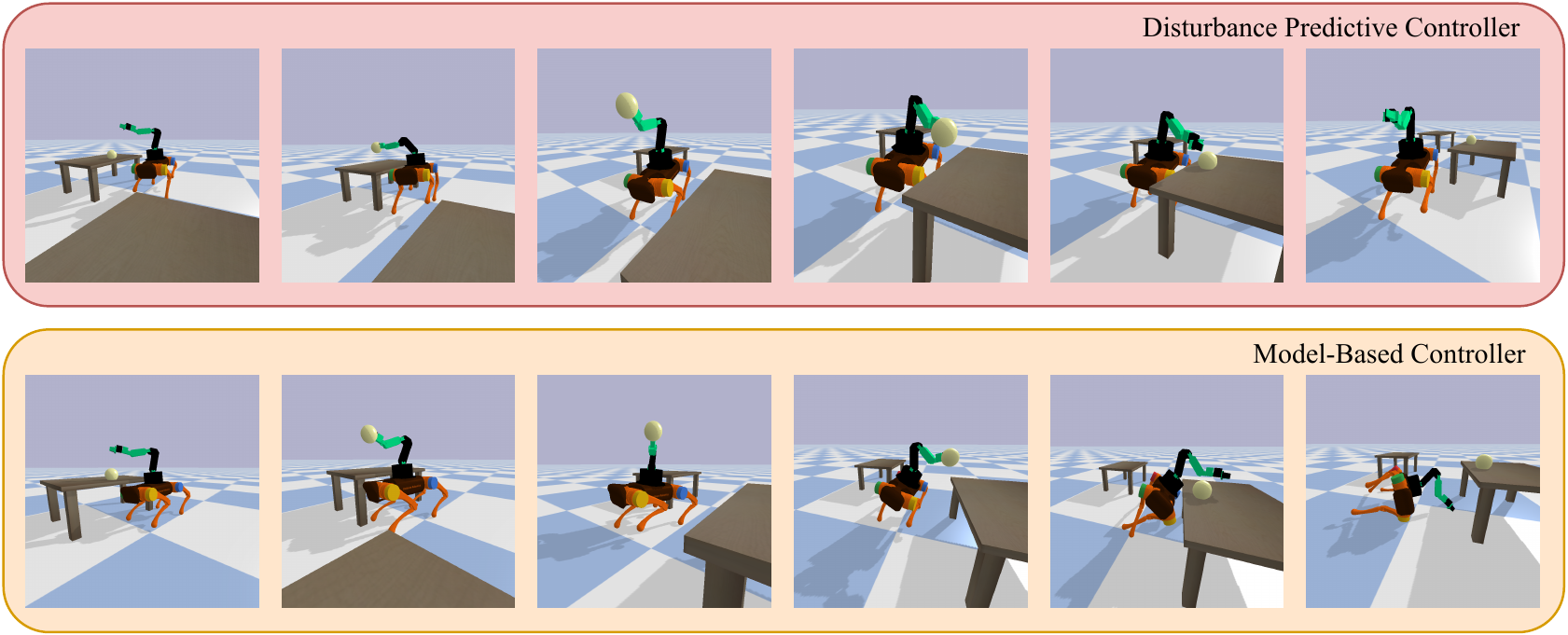}
    \includegraphics[width=0.4\textwidth,height=0.4\textheight]{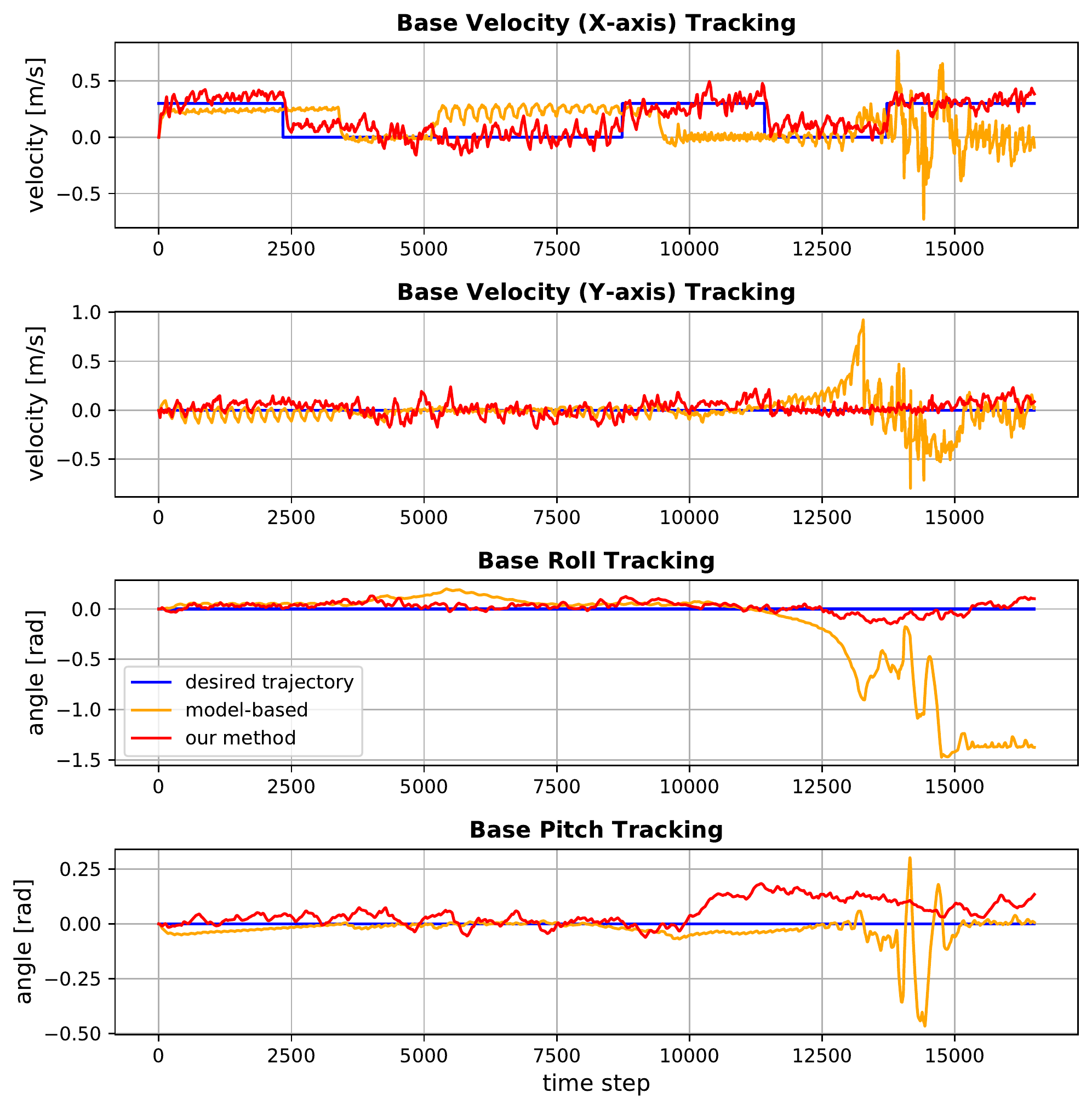}
    \caption{Trajectory tracking of DPC and MBC on the ball-lifting task in simulated environments. DPC has better stability when the arm is loaded.}
    \label{Fig.lifting}
    \end{figure}
    The high-level estimator and latent dynamic adapter can be used directly in different tasks by the same manipulator without fine-tuning.
    The ball lifting experiment is visualized to demonstrate the feasibility of our method, and the experimental results are shown in Fig. \ref{Fig.lifting}. In this task, the robot needs to keep its body moving steadily along the desired trajectory with a 0.3 kg ball. We compare the disturbance predictive control(DPC) with a model-based controller(MBC)\cite{di2018dynamic}. 
    The results indicate that the robotic arm's disturbance can cause the robot to lose balance and that DPC can assist the robot to maintain balance when the robotic arm is interacting.


    

    \begin{table}[t]
    \centering
    \caption{Effects of migration to different tasks.
    }\label{tab}
    \begin{tabular}{lllll}
    \toprule
    Arm Type             & MBC      & DPC      & biceps length(m)      & gripper weight(kg)\\ \midrule
    regular              & -969.25 & 401.13 & 0.14 & 0.1  \\
    longer               & -491.92  & 402.54  & 0.28  & 0.1   \\
    heavier & 331.25  & 351.88 & 0.14  & 0.5   \\
    double           & -729.48 & 383.15  & 0.14 & 0.1 \\ 
    \bottomrule
    \end{tabular}
    \end{table}
    We use different robotic arms to perform corresponding tasks with the trained policy to further test the network's generalization capabilities. To evaluate the migration capabilities of our method, we used various lengths and weights, as well as different numbers of robotic arms.
    All migration processes need to collect random motion samples of the different robotic arms to train the corresponding encoder. The $z_t$ processed by the different encoders has similar properties for the agent due to the restrictions of the original decoder. Therefore, the trained policies and decoders can be used directly without retraining.
    We tested the pushing task and two types of carrying tasks that are not suitable for the original robotic arm because of the shape, length and quantity. Corresponding results and parameters of the manipulator are listed in Table \ref{tab}.
    Experiments show that DPC can resist the disturbance of manipulators.
    This method has higher sample efficiency. We use 3e4 samples to complete the migration, while the original model training requires 1e6 samples.  
    \begin{figure}[htbp]
    \centering
    \includegraphics[width=0.45\textwidth]{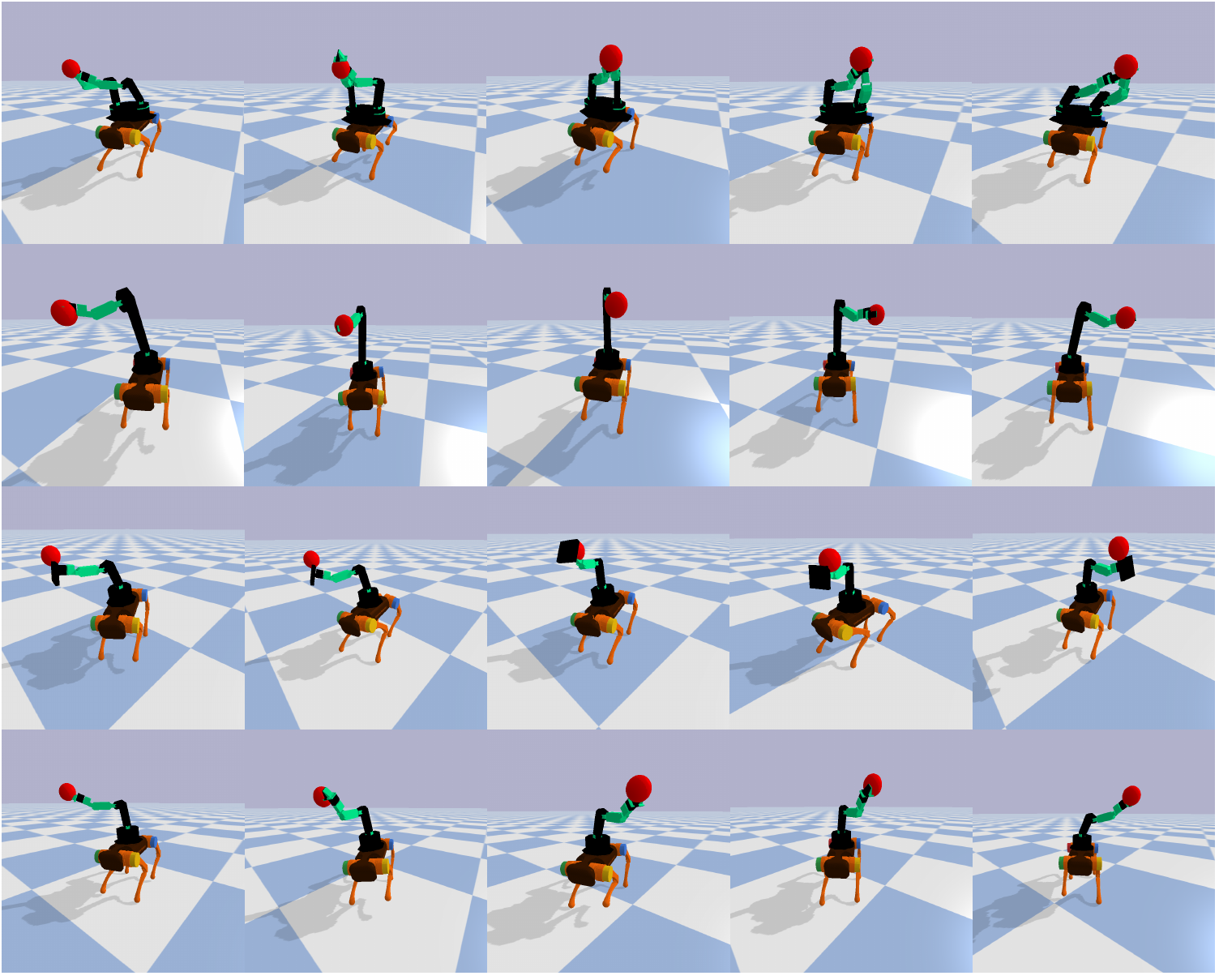}
    \caption{Different adapted robotic arms on the task of reaching in simulation. Different lengths, weights and quantity of robotic arms were used to test the migration capabilities of our model.}
    \label{Fig.arms}
    \end{figure}
    
    \begin{figure*}[htbp]
    \centering
    \includegraphics[width=0.95\textwidth]{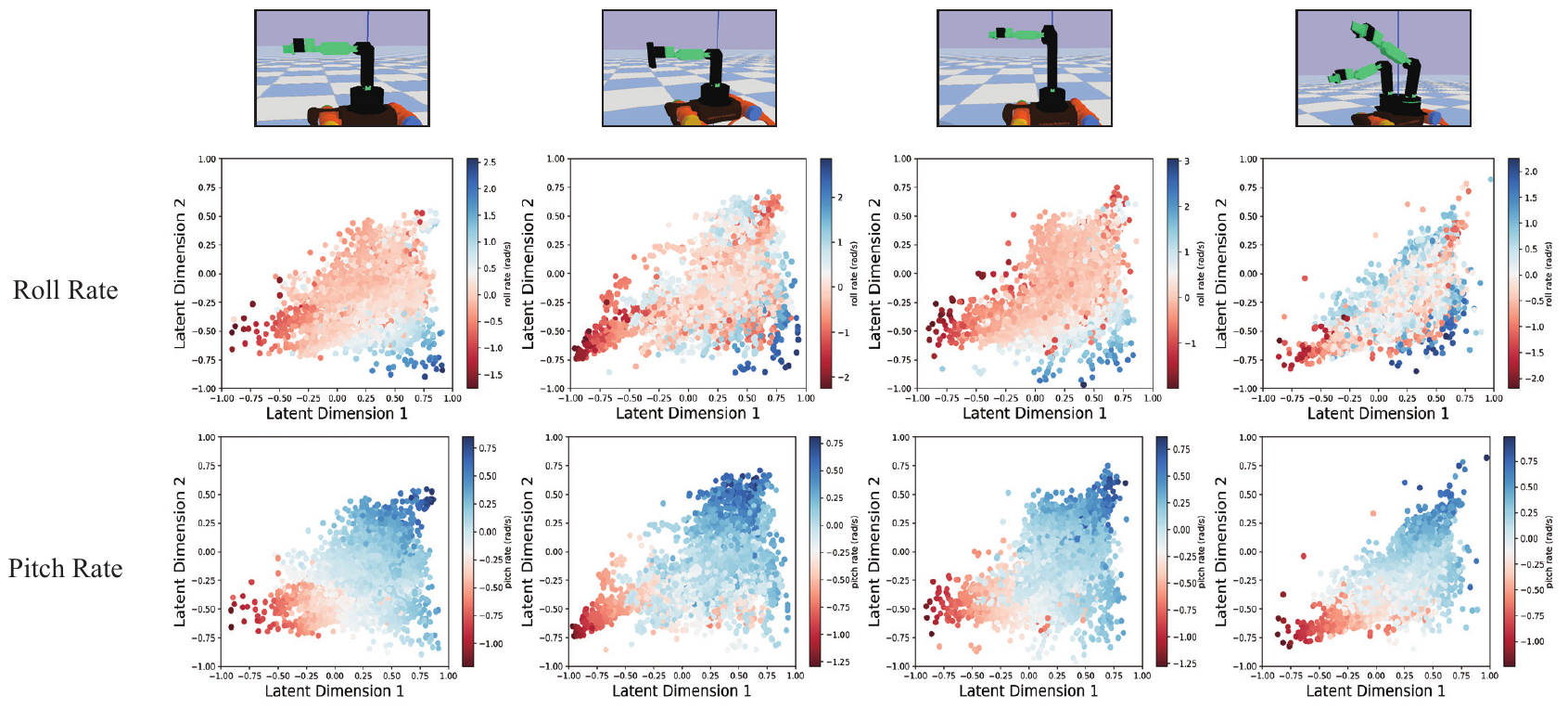}
    \caption{Comparison of the latent state with different robotic arm.}
    \label{Fig.result_z}
    \end{figure*}
    \begin{figure*}[htbp]
    \centering
    \includegraphics[width=0.85\textwidth]{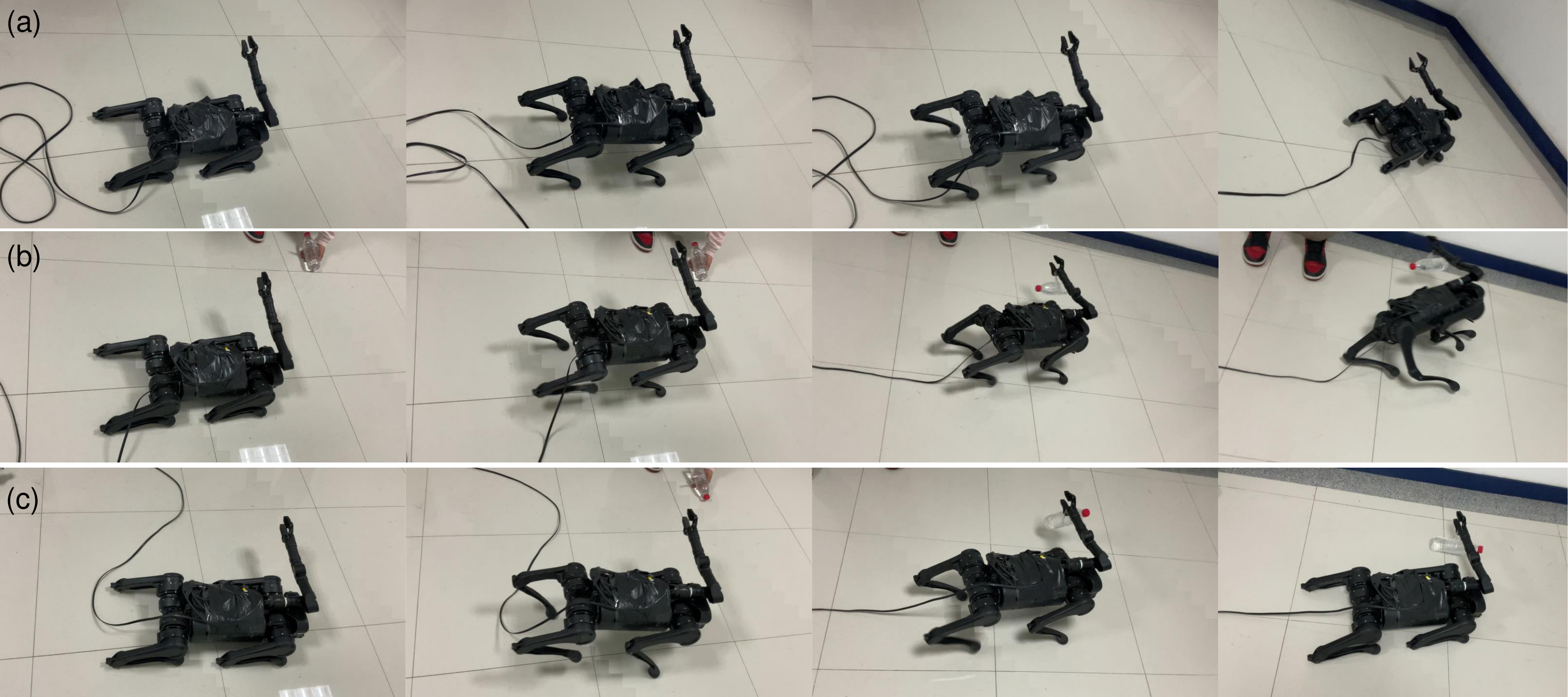}
    \caption{The real experiments screenshots of legged mobile manipulation. (a) no payload; (b) MBC with payload; (c) DPC with payload.}
    \label{Fig.real_exp}
    \end{figure*}
    We finally test the case of carrying multiple robotic arms. Experiments demonstrate the potential of our model for legged mobile manipulation as shown in Fig. \ref{Fig.arms}.
    As far as we know, this is the first time that dual robotic arms have been used on a quadruped robot.

    To demonstrate the relationship between the latent state $z_t$ and different robotic arms, we visualize the latent state $z_t$ of various encoders and $s_{t+1}^{drp}$ of the body as shown in Fig. \ref{Fig.result_z}. The image illustrates that the latent state from different encoders of diverse robotic arms have a similar feature, which explains why models can be migrated.
    
    We further migrate our algorithm to a different type of robotic arms and use them in a real robot to test the mobility of our method and its ability in the real robot, 
    We use the trained decoder with a Unitree A1 legged robot and WidowX robotic arm to supervise the encoder of a different structure of the manipulator and deploy our method on a real legged manipulator system that includes a Unitree A1 legged robot and BlueprintLab Reach5Mini arm, as shown in Fig. \ref{Fig.real_exp}. The Fig. \ref{Fig.real_exp}(a) shows the basic walking without any external disturbance and the robot can finish the task successfully. Fig. \ref{Fig.real_exp}(b) is the baseline of MBC, which fails to walk when adding a payload. Fig. \ref{Fig.real_exp}(c) demonstrate the successful walking with the DPC method, even with an unknown payload and random disturbance from the swinging payload. 
    The sim2real experiments show the generalization and effectiveness when the legged mobile manipulator has an unknown disturbance.


\section{Conclusion}
\label{sec:conclusion}
    We propose a hierarchical framework that merges an disturbance-based control with the reinforcement learning and the forward model. A forward model is used to learn the dynamic representations of different robotic arms for quick migration. We test our framework in challenging scenarios with different external disturbances, the results are visualized to illustrate the capabilities of the forward model and the reinforcement learning to adapt to the different tasks. The potential of our model is further demonstrated by the rapid migration of different types and numbers of robotic arms based on a few random motion samples. We believe this is a step forward in combining the advantages of model-based control and reinforcement learning.
    Further work will incorporate visual information for end-to-end legged mobile manipulation tasks.







\bibliographystyle{IEEEtran}
\bibliography{reference}
\end{document}